\documentclass[10pt,letterpaper]{article}

\usepackage{graphicx}
\usepackage{amsmath}
\usepackage{amssymb}
\usepackage{booktabs}

\usepackage{algorithm}
\usepackage{algcompatible}
\usepackage{amsmath}
\usepackage{booktabs}

\usepackage{xcolor}

\usepackage[pagebackref,breaklinks,colorlinks]{hyperref}

\usepackage[capitalize]{cleveref}
\crefname{section}{Sec.}{Secs.}
\Crefname{section}{Section}{Sections}
\Crefname{table}{Table}{Tables}
\crefname{table}{Tab.}{Tabs.}

\begin{document}
\date{}


\title{Theoretical Corrections and the Leveraging of Reinforcement Learning to Enhance Triangle Attack}

\author{Nicole Meng\\
University of Connecticut\\
{\tt\small nicole.meng@uconn.edu}
\and
Caleb Manicke\\
University of Connecticut\\
{\tt\small caleb.manicke@uconn.edu}
\and
David Chen\\
University of Connecticut\\
{\tt\small david.y.chen@uconn.edu}
\and
Yingjie Lao\\
Tufts University\\
{\tt\small yingjie.lao@tufts.edu}
\and
Caiwen Ding\\
University of Minnesota\\
{\tt\small dingc@umn.edu}
\and
Pengyu Hong\\
Brandeis University\\
{\tt\small pengyu.hong@brandeis.edu}
\and
Kaleel Mahmood\\
University of Rhode Island\\
{\tt\small kaleel.mahmood@uri.edu}
}
\maketitle

\begin{abstract}
Adversarial examples represent a serious issue for the application of machine learning models in many sensitive domains. For generating adversarial examples, decision based black-box attacks are one of the most practical techniques as they only require query access to the model. One of the most recently proposed state-of-the-art decision based black-box attacks is Triangle Attack (TA). In this paper, we offer a high-level description of TA and explain potential theoretical limitations. We then propose a new decision based black-box attack, Triangle Attack with Reinforcement Learning (TARL). Our new attack addresses the limits of TA by leveraging reinforcement learning. This creates an attack that can achieve similar, if not better, attack accuracy than TA with half as many queries on state-of-the-art classifiers and defenses across ImageNet and CIFAR-10.





\end{abstract}


\section{Introduction}

As machine learning models advance into real-world applications, security concerns around these models are increasing. Identifying and addressing the vulnerabilities of machine learning models is crucial to ensuring their safety. In recent years, researchers have developed effective and efficient attacks that pose greater adversarial threats, along with robust defenses to protect these models. This has led to the emergence of adversarial machine learning.



\subsection{Threat Model}

Based on the attacker's knowledge perspective, existing attacks can be broadly categorized into black-box and white-box attacks \cite{carlini2017towards}. White-box attacks assume the attacker has full access to the target victim model, such as model parameters or the training data set \cite{lin2021ml, mahmood2021robustness}. In contrast, black-box attacks are implemented without prior knowledge of the target victim model, except for the model output. Out of existing black-box attacks, decision-based attacks are particularly significant because they require the minimum information (final output label) for a successful attack, making them the most applicable, and therefore most threating to real-life deep neural networks \cite{lin2021ml, mahmood2021back}. 


In this work, we explore the black-box threat model, where we do not have access to model gradients. Our threat is decision-based and works in the $l_2$ norm. The reasoning and efficiency of our threat model is explained in the following sections. 

\subsection{Black-box Attacks}
Black-box attacks can be broadly categorized into three categories: Transfer Attack, Score-based Black-box Attack, and Decision-based Attack\cite{lin2021ml}. 
\begin{itemize}
  \item \textbf{Transfer Attack:} This type of attack utilizes a model trained by the attacker (often using part of the original training data) to generate adversarial examples; these examples are crafted in such a way that they are also likely to be effective against the target model due to similarities in model behavior \cite{goodfellow2014explaining, kurakin2018adversarial}. 
  \item \textbf{Score-based Black-box Attack:} In score-based attacks, the attacker queries the target model with inputs and utilizes the returned scores (such as class probabilities) to iteratively modify the inputs and craft adversarial examples without needing direct access to the model’s architecture or weights \cite{papernot2017practical}. Examples of such attack includes ZOO Attack, which aims to find adversarial examples based on zero-order-optimization \cite{chen2017zoo}, and Square Attack, which utilizes random search \cite{Kim2021RobustDB}, is resistant to gradient masking, and further focus on query efficiency\cite{andriushchenko2020square}.  
  \item \textbf{Decision-based Attack:} This category involves generating adversarial examples by only having access to the final decision or classification of the target model (like class labels), making it a more challenging but highly realistic scenario where the attacker tweaks inputs based only on the output decisions \cite{brendel2018decision, mahmood2021back}.
\end{itemize}

The decision-based attack is unparalleled among black-box attacks, as it relies solely on the final decision class label of the model to generate a successful attack \cite{brendel2018decision}. Decision-based attacks hold high relevance to real-world scenarios, rendering them highly persuasive for rigorously testing the robustness of machine learning models and networks \cite{NEURIPS2022_544696ef}. Furthermore, its immunity to standard defense mechanisms sets it apart from other attack types. Consequently, the development of more resilient decision-based attack methods is not only pivotal but also pragmatic in comprehensively evaluating the safety and reliability of machine learning. The first successful method in decision-base attacks is BoundaryAttack \cite{brendel2018decision}, which generates a large perturbation on the victim image first, then efforts are made to perform random walks on the decision boundary to minimize perturbation. Other successful decision-based attacks include SignOPT \cite{cheng2019sign}, HopSkipJumpAttack (HKJA) \cite{chen2020hopskipjumpattack}, and QEBA \cite{li2020qeba}. 

Despite the impressive attack success rate of these methods, they all face a common issue: query-efficiency. Later attack methods have made an effort to improve the query efficiency of decision-based attacks, including qFool \cite{liu2019geometry}, GeoDA (Geometry-Based Decision-Adaptive attack) \cite{rahmati2020geoda}, and Surfree \cite{maho2021surfree}. Most of the existing decision-based attacks first start by generating a large perturbation, then optimize the perturbation by restricting the adversarial example at each iteration on the decision boundary. They usually use different algorithms to estimate gradient and the boundary, causing the limitations on query-efficiency. Triangle Attack (TA) \cite{wang2022triangle} was proposed in 2022 and outperformed all the mentioned decision-based attacks proposed prior to it. TA is extremely query-efficient, achieving a much higher success rate within 1,000 queries on ImageNet models, and requiring fewer queries than existing decision-based attacks to reach the same success rate across various perturbation budgets \cite{wang2022triangle}. Since the inception of Triangle Attack, attacks such as DEAL \cite{shen2023decision} and CGBA in 2023 \cite{reza2023cgba}, BounceAttack \cite{wan2024bounceattack} and QE-DBA in 2024 \cite{zhang2024qe}  have found success in generating adversarial examples with smaller perturbations than prior attacks in both the $l_2$ and $l_\infty$ domain.

\subsection{Main Contribution}
Triangle Attack is the first approach that leverages geometric information to directly optimize perturbations in frequency space without imposing constraints on the adversary's location with respect to the decision boundary. Empirically, this technique achieves high query efficiency, making it the most query-efficient successful attack in the field. Extensive experiments showed that TA exhibits a much higher attack success rate than existing state-of-the-art within 1,000 queries and needs a much smaller number of queries to achieve the same attack success rate with the same perturbation. To improve Triangle Attack's performance, we evaluated our method, TARL, using the same dataset, model parameters, and evaluation metrics. Our main contributions can be summarized as follows:

\begin{enumerate}
    \vspace{-0.15 cm}
    
  \item TARL directly addresses the limitations of the most state-of-the-art decision-based attack, Triangle Attack, making it achieve similar or more optimal accuracy with less queries.

  \vspace{-0.25 cm}
  \item TA only showed performance on 5 models on only 200 images from ImageNet, whereas we compare performance of both TA and TARL on 9 models across two datasets, ImageNet and CIFAR-10.

  \vspace{-0.25 cm}
  \item TA did not show performance against defenses; we compare performance of both TA and TARL against the Diffusion model, a SOTA defense.

  \vspace{-0.25 cm}
  \item Decision-based models have managed to generate successful adversarial examples with minimal perturbations on a $<$1000 query budget on CIFAR-10 \cite{shen2023decision, zhang2024qe} and on a $>$1000 query budget on ImageNet \cite{reza2023cgba, wan2024bounceattack}. TARL generates adversarial examples on a 500 query budget that perform similarly to examples produced on a 1000 query budget across most popular decision-based attacks.
\end{enumerate}

\section{Triangle Attack Limitations}
Triangle Attack (TA) \cite{wang2022triangle} is one of the most SOTA query-efficient decision-based attack algorithms. In order to address the limitations of TA, a brief overview of the TA algorithm is shown in the following subsection. 

\subsection{Triangle Attack Methodology}

\begin{figure}
  \centering
  \includegraphics[scale=0.5, width=7cm]{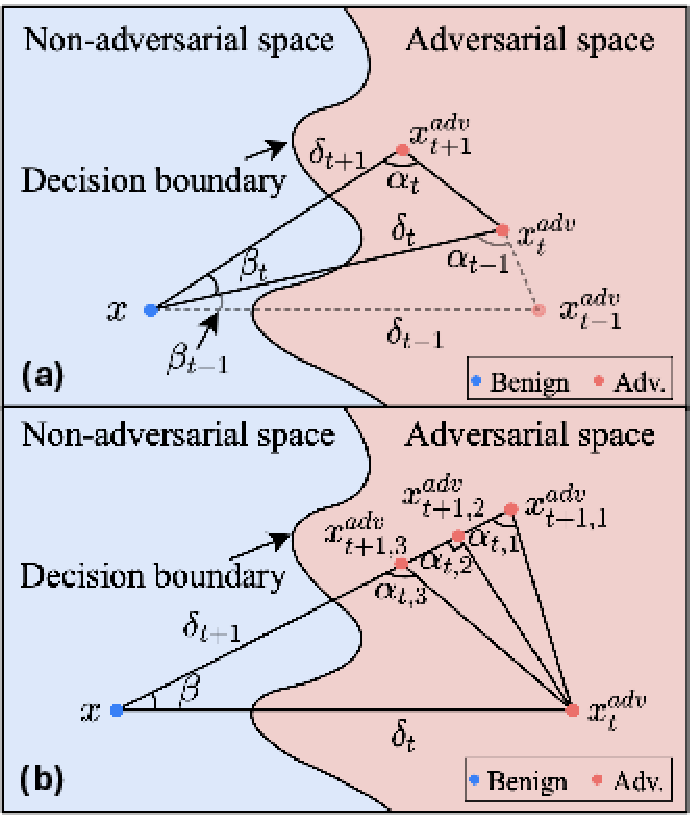}
  \caption{The Effect of the magnitude of $\alpha$ \cite{wang2022triangle}}
  \label{fig:normalTA}
\end{figure}

Since the TA attack relies on iteratively exploring the 2D subspace of images, TA's algorithm starts by applying the Discrete Cosine Transform (DCT) algorithm to the input image to extract it from the original image space and reduce its dimensionality~\cite{ahmed1974discrete}. Afterwards, TA starts applying a large perturbation to the target image, then uses geometric information between adversarial candidates for optimization. As demonstrated in Figure ~\ref{fig:normalTA} Part (a), at the $t$-th arbitrary iteration of the triangle attack, the benign image, current adversarial example, and the next candidate adversarial example at the $(t + 1)$-th iteration $x_{t+1}^{adv}$ forms a triangle for any iterative attacks. To generate $x_{t+1}^{adv}$ with a smaller perturbation than $x_{t}^{adv}$ at the $(t+1)$-th iteration, the Euclidean distance between $x_{t+1}^{adv}$ and $x$ need to be smaller than the Euclidean distance between $x_{t}^{adv}$ and $x$, i.e., $\delta_{t+1}$ $<$ $\delta_t$.  


Based on this property, the TA algorithm can be summarized as follows after obtaining the low frequency space of the victim image: First, a directional line across the benign sample is randomly selected to construct a 2-D subspace, where the triangles with three vertices $x$, $x_{t}^{adv}$, and $x_{t+1}^{adv}$, are iteratively constructed with two adjustable angles, searched angle $\beta$ and learned angle $\alpha$. at the $t$-th iteration. The searched angle $\beta$ and learned angle $\alpha$ are further updated and adjusted to achieve perturbation optimization - that is, to find the adversarial image with the minimal possible perturbation within the given budget - for each constructed triangle. Triangle attack is the first work that directly optimizes the perturbation in the frequency space via geometric information without restricting the adversarial image to the decision boundary or estimating gradients at each iteration. This allows TA to be extremely query-efficient while achieving a much higher success rate than other SOTA black-box attacks.

\subsection{Triangle Attack Limitation}
\label{sec:limitation}

However, there is a major limitation on the update method of the learned angle $\alpha$ of the algorithm of Triangle Attack. According to Proposition 1 in the paper \cite{wang2022triangle}, “With the same angle $\beta$, a smaller angle $\alpha$ makes it easier to find an adversarial example while a larger angle $\alpha$ leads to smaller perturbation.” We will consider scenarios where this objective can lead to TA failing to produce successful adversarial examples.

Figure ~\ref{fig:normalTA} Part (b) illustrates a scenario where TA works. With the same $\beta$, the adversarial examples with learned angle ${\alpha}_{t,3}$ is closer to the boundary with the smallest perturbation while the adversary example with ${\alpha}_{t,1}$ is further away from the benign image and decision boundary, making it easier to secure an adversarial candidate. Therefore, TA adaptively adjusts the angle $\alpha$: at the $t$-th iteration, if the image is adversarial, $\alpha$ is increased by a change rate $\gamma$ in hope of a better adversarial image at the next iteration. Conversely, suppose the image is not adversarial at the $t$-th iteration, in which case $\alpha$ is decreased by change rate times a constant, $\lambda\gamma$, such that the next candidate will be more likely to be adversarial. 

\begin{figure}
  \centering
  \includegraphics[scale=0.5, width=7cm]{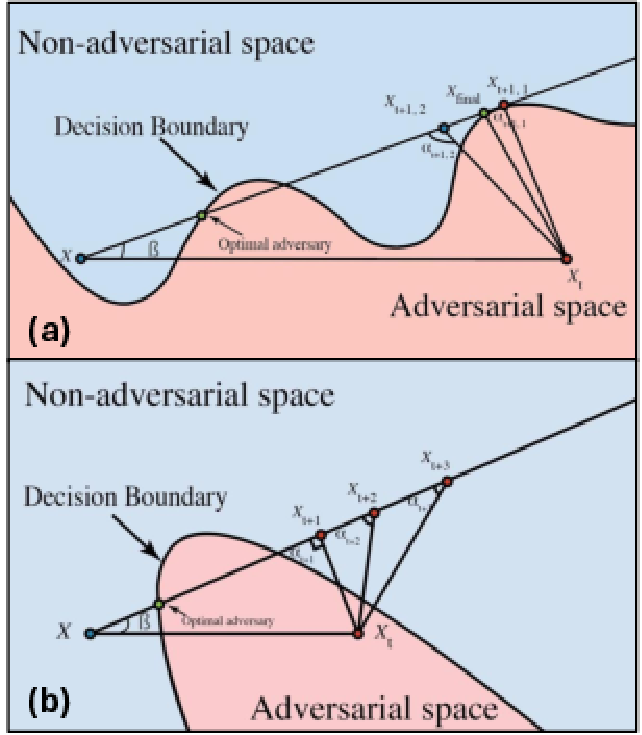}
  \caption{Illustration of cases where TA does not sucessfully generate an adversarial example}
  \label{fig:edgeCases}
\end{figure}

Though the alpha-update algorithm and proposition for the learned angle $\alpha$ seem to be fair in the figure illustration, they only guarantee the most optimal results at some times. We show this through Figure~\ref{fig:edgeCases} Part (a); here, TA fails. With the fixed $\beta$ and the decision boundary illustrated in Figure~\ref{fig:normalTA} Part (b), the alpha update algorithm given by TA will start the search process with ${\alpha}_{t+1, 1}$. Following the algorithm, since $x_{t+1, 1}$ is adversarial, alpha is increased to ${\alpha}_{t+1, 2}$, without loss of generality. The resulting image  $x_{t+1, 2}$ is not adversarial, causing alpha to decrease. Eventually, when the maximum number of queries is reached, the green point $x_{final}$ is reached and returned as the best adversary found by TA. However, it is obvious that the actual optimal adversarial example is different from what the alpha update algorithm of TA returns, as the algorithm is based on an inaccurate proposition. 

In extreme cases, such as the one shown in Figure ~\ref{fig:edgeCases} Part (b), the alpha update algorithm might never be able to find an adversarial example when there are possible adversarial candidates present. TA fails to find an adversarial example starting from ${\alpha}_{t+1}$. As alpha keeps decreasing, the algorithm will use up all the queries and the attack will fail when adversarial examples can be found if alpha is increased instead of decreased. 

According to the experimental results of TA, it is the most query-efficient decision-based method compared to other attacks. However, the alpha update algorithm shows there is potential for improvement.





\subsection{Triangle Attack Using Reinforcement Learning (TARL)}

Reinforcement learning is a sub-field of machine learning that focuses on developing algorithms and models that enable an agent to learn optimal actions in an environment by interacting with it \cite{kaelbling1996reinforcement, mnih2015human}. The agent learns through trial and error, receiving feedback based on its actions. The goal of reinforcement learning is to find an optimal policy that maximizes the cumulative reward over time using historical data \cite{kaelbling1996reinforcement}. Q-learning is a popular algorithm used in reinforcement learning to learn an optimal approach for decision-making \cite{watkins1989learning}. It is a model-free, value-based method that uses a Q-table to store the expected cumulative rewards for each possible action in each possible state. The Q-value represents the expected cumulative reward the agent can obtain by taking a certain action from a particular state and following the optimal policy thereafter \cite{melo2001convergence, watkins1992q}. 

To address the problems with the current Triangle Attack method outlined in Section~\ref{sec:limitation}, 
we revised the alpha update algorithm and introduce TARL. TARL implement a Q-learning algorithm to train the agent to find the optimal value of alpha based on historical observation data on alphas, corresponding $l_2$ norm values, and boolean variables indicating whether the attack is successful. With every query used on a candidate adversary, we record the three data entries and add them to historical data to train our agent. TARL will allow the agent to find the most optimal parameter (alpha) based on the characteristics of different shapes of decision boundaries, therefore producing a more accurate, optimal adversarial example using fewer queries. Overall, our Q-learning algorithm integrates with the Triangle Attack method by training an agent to find the optimal value of alpha based on historical data and leveraging its learning capabilities to adjust the parameter value adaptively. This approach significantly improves the effectiveness of the attack and enhances the overall query performance of the Triangle Attack method according to experimental results.

\section{Methodology}

In this section, we will first introduce preliminaries. Then we will introduce a detailed overview of the methodology of our proposed method with reinforcement Q-learning.

\subsection{Preliminaries}
Given a classifier $f$ with parameters $\theta$ and a benign sample $x\in X$ with ground-truth label $y\in Y$, where $X$ represents the set of all input image space and $Y$ denotes the output space, the goal of adversarial attacks is to find an adversarial example $x^{adv}\in X$ to cause a misclassification of the victim classifier model $f$ such that the equation ~\ref{eq:findAdv} holds:
\begin{align}
    f(x^{adv}; \theta) \neq f(x, \theta) = y\quad s.t.\quad \| x^{adv} - x \|_{p} < \epsilon
    \label{eq:findAdv}
\end{align}
Note that $\epsilon$ is the perturbation budget that ideally will be minimized after generating the first adversarial example with a large perturbation, as demonstrated in equation ~\ref{eq:minPerturb}. 
\begin{align}
    min\| \delta \|_{p} \quad s.t. \quad f(x+\delta; \theta) \neq f(x, \theta) = y
    \label{eq:minPerturb}
\end{align}

\subsection{Revised Triangle Attack Using Reinforcement Learning}
Along with most other decision-based attack methods, TARL is conducted based on an assumption for any victim deep neural classifier model $f$: \\\\
\textbf{Assumption 1. }Given a benign sample $x$ and a perturbation budget $\epsilon$, 
there exists an adversarial example with perturbation $\| \delta \|_{p} < \epsilon$
towards the decision boundary which can mislead the target classifier $f$. \\


\begin{algorithm}
\caption{Triangle Attack with Reinforcement Learning}
\label{TARL}
\textbf{Input}: Target classifier $f$ with parameters $\theta$; 
Benign sample $x$ with ground-truth label $y$; Maximum number of queries $Q$; Maximum number of iteration $N$ for each sampled subspace; 
Lower bound $\underline{\beta}$ for $\beta$; Learning rate $lr$; Discount factor $\gamma$; Exploration rate $\epsilon$; Current state/alpha value $S^{'}$.

\textbf{Output}: An adversarial example $x^{adv}$

\begin{algorithmic}[1]
  \STATE Initialize perturbation $\delta_0$; a Q-table with Alg. 2;
  \WHILE{$q < Q$}
    \STATE Sampling 2-D subspace $S_t$
    \STATE $\beta_{t,0} = \max(\pi - 2\alpha, \beta)$;\;
    \IF{$f(T(x, x_{t}^{adv}, \alpha_{t,0}, \beta_{t,0}, S_{t});\theta) = f(x;\theta)$}
        \STATE $q = q + 1$;\;
        \IF{q-table is empty}
                \STATE Set $\alpha_{t,0}$ as the starting state;
                \STATE Suggest next alpha value $\alpha_{t,0}$ with Alg. 3;
                \STATE Update current alpha and state;
                \STATE Go to line 17;
        \ELSE
                \STATE Store $\| {x}_{t}^{adv} - {x} \|_2$ , the label of the current state;
                \STATE Calculate reward, update q\_table using Alg. 4;
                \STATE Update current alpha and state;
        \ENDIF
        \IF{$f(T(x, x_{t}^{adv}, \alpha_{t,0}, -\beta_{t,0}, S_{t});\theta)=f(x;\theta)$}
        \STATE $q=q+1$;
        \STATE Store $\| {x}_{t}^{adv} - {x} \|_2$ , the label of the current state;
        \STATE Calculate reward, update q\_table using Alg. 4;
        \STATE Update current alpha and state;
        \STATE Go to line 3;
        \ENDIF
\ENDIF
\STATE $\bar{\beta}_{t,0} = min({\pi/2, \pi - \alpha})$;
\FOR{$i \gets 1$ to $N$}
\STATE $\beta_{t, i+1} = (\bar{\beta}_{t, i}, {\beta}_{t, i})/2$;
\IF{$f(T(x, x_{t}^{adv}, \alpha_{t,0}, \beta_{t,0}, S_{t});\theta) = f(x;\theta)$}
\STATE $q = q+1$;
\STATE Store $\| {x}_{t}^{adv} - {x} \|_2$, the label of the current state;
\STATE Calculate reward, update q\_table using Alg. 4;
\STATE Update current alpha and state;

\IF{$f(T(x, x_{t}^{adv}, \alpha_{t,0}, -\beta_{t,0}, S_{t});\theta) = f(x;\theta)$}
\STATE $\bar{\beta}_{t, i+1} = \beta_{t, i+1}$;
\STATE $\beta_{t,i+1} = \beta{t,i}$;
\ENDIF
\ENDIF
\STATE $q=q+1$;
\STATE Store $\| {x}_{t}^{adv} - {x} \|_2$ and the label of the current state;
\STATE Calculate reward, update q\_table using Alg. 4;
\STATE Update current alpha and state;
\ENDFOR
\ENDWHILE
\STATE \textbf{return} $x_{t}^{adv}$;
\end{algorithmic}
\end{algorithm}

The assumption is essentially saying that the attack is feasible within the given budget on the victim classification model, which has been proven true by previous research and studies \cite{athalye2018obfuscated, brendel2018decision, carlini2017towards, wang2021boosting}. If the assumption does not hold for a model $f$ and the given perturbation budget, then $f$ is considered to be well-defended and secure, therefore removing $f$ from the discussion and the scope of the attack \cite{wang2022triangle}. Under this assumption, TARL generates the starting adversarial example with a large perturbation using previously established methods and algorithms. As outlined in \ref{TARL}, TARL followed TA's way of generating the first adversarial example, which is through a pre-implemented Binary Search \cite{li2020qeba, maho2021surfree, rahmati2020geoda}. After generating the first adversarial example, the perturbation optimization process is done in a 2-D subspace sampled from the frequency space of the image though the Discrete Cosine Transform(DCT) \cite{wang2022triangle}. TARL kept the transformation and the subspace construction steps identical to TA as it has been proven effective and contains the most critical information of the image \cite{wang2022triangle}. 

After securing the 2-D subspace to construct candidate triangles on, the searched angle $\beta$ need to be determined before adjusting the learned angle $alpha$. Following the steps of TA, binary search is adopted to find the optimal $\beta^* \in [max (\pi-2\alpha, \beta_0), min(\pi-\alpha, \pi/2)]$, which should be as large as possible to minimize perturbation based on the following proposition \cite{wang2022triangle}:With the same learned angle $\alpha$, a larger $\beta$ will lead the resulting image to be closer to the benign image, i.e. the larger $\beta$ is, the smaller the perturbation is. 

\begin{algorithm}
\caption{Construct Q-table}
\label{Construct}
\textbf{Input}: Number of possible actions $A$; The lower/upper bound of alpha $\alpha_{min}, $$\alpha_{max}$; Step size for alpha $step$. 

\textbf{Output}: An empty Q-table;

\begin{algorithmic}[1]
\STATE $S= 0$;
\STATE $\alpha = \alpha_{min}$;
  \WHILE{$\alpha < \alpha_{max}$}
    \STATE $S = S+1$;
    \STATE $\alpha = \alpha + step$;
\ENDWHILE
\STATE q\_table = 2-D array with S rows and A columns filled with 0 value;
\STATE \textbf{return} q\_table;
\end{algorithmic}
\end{algorithm}

\begin{algorithm}
\caption {Epsilon-greedy strategy to suggest next update alpha}
\label{Epsilon}

\textbf{Input}: Current state/alpha value $S'$; Current value of alpha $\alpha$, Q-table $q\_table$; An array of actions $actions$ with length equals to the number of possible actions $A$. 

\textbf{Output}:The action taken and the updated state, the current state will be updated. 

\begin{algorithmic}[1]

\IF {$p < \epsilon$} 
    \STATE next $\alpha$ = take $action[0]$;
    \STATE save the action taken as $A'$;
\ELSE
    \STATE next $\alpha$ = take the action with the highest Q-value in $q\_table$ given $S'$;
    \STATE save the action taken as $A'$;
\ENDIF
\STATE Clip next $\alpha$ value within the proper range between $\alpha_{min}$ and $\alpha_{max}$;
\STATE Update $S'$ and $\alpha$ with the next set of values;
\STATE \textbf{return} $A'$, $S'$;
    
\end{algorithmic}
\end{algorithm}

\begin{algorithm}
\caption {Q-table Update}
\label{Update}

\textbf{Input}: Current state/alpha value $S'$; Corresponding $l2$ value, Corresponding class label $y'$, Q-table $q_table$; Learning rate $lr$; Discount factor $\gamma$; Exploration rate $e$.

\textbf{Output}: None, the Q-table will be updated. 

\begin{algorithmic}[1]
\IF{the current image is adversarial}
\STATE reward = $-l_2$;
\ELSE
\STATE reward = 0;
\ENDIF
\STATE Update $q\_table$ based on the Q-learning update rule;

\end{algorithmic}
\end{algorithm}

\textbf{Adjusting the learned angle $\alpha$ using reinforcement learning.}
TARL implements a Q-learning algorithm to adaptively adjust alpha based on historical data of the queries used. \cite{watkins1992q}. This will allow the agent to explore different alpha values and learn the characteristics of different boundary shapes. Additionally, it can handle continuous action spaces, which is suitable for the learned angle alpha in some predefined range. Furthermore, TARL is more generalizable than TA’s alpha update method as, rather than having a fixed update algorithm that we have proven to be not generalizable to all shapes of decision boundaries, the agent can learn the unique characteristic of each subspace and decision boundary. Finally, TARL is able to take into consideration previously explored alpha values and corresponding $l_2$ distances when suggesting the next alpha, making it more query-efficient, since our Algorithm \ref{Epsilon} can converge to the optimal point with fewer steps, considering that each step is an educated guess.

To calculate and store the expected cumulative rewards (known as Q-values) for each possible action in each state, TARL uses a Q-table, as suggested by the Q-learning Algorithm \cite{watkins1992q}. Algorithm \ref{Construct} builds a Q-table where the size equals the number of state-action pairs of the Q-learning model. In this case, the states represent all possible different values of alpha and the actions represent whether to increase or decrease alpha. Since the optimization goal is to minimize the $l_2$ value, the reward is set to be equal to the $\frac{1}{l_2}$ value if the resulting image is adversarial, and to $0$ if it is not at each iteration. As TARL train the agent, alpha values, their corresponding $l_2$ values, and the resulting labels are recorded so that we can calculate and store the corresponding Q-values. This Q-table becomes a reference table for our agent to select the best action based on the Q-values, where the Q-values are calculated through the Q-learning update rule \ref{Update}, a common and effective approach in updating the Q-table \cite{watkins1992q}.

\section{Experiments}
Since TARL aims to improve the performance of Triangle Attack, to ensure a controlled environment, the experiments are designed to maintain consistency in the parameters, experimental dataset, and victim model used in TA's experiments. Furthermore, since TA only utilizes 5 vanilla victim models and 200 images from ImageNet, we aim to further extend the scope of the experiments using 2 full datasets, 10 victim models, and 1 defense model. Both the Vision Mamba and Diffusion defense model are new models. These settings allow us to better highlight the efficiency of both TA and TARL on a larger scale. 


\subsection{Experimental Setup}

\textbf{Models} We included the 5 models utilized in TA:VGG-16 \cite{VGG}, Inception-v3 \cite{BatchNorm}, ResNet-18, ResNet-101 \cite{ResNet}, and DenseNet-121 \cite{DenseNet}. We also included larger models such as the ViT-L \cite{BERT}, ViT-B \cite{BEiT}, BiT-M \cite{BiT}, and Mamba \cite{zhu2024vision}. This range from early convolutional models to vision transformers allows our experiment to evaluate both TA and TARL on an encompassing scale of model complexity.

The inclusion of Vision Mamba allows us to see how a state-of-the-art state-space model would perform against our attack. Mamba is a new model that achieves better computational efficiency than transformers by using selective state-space (SSM) blocks to compress data, rather than relying on the traditional self-attention mechanisms that process the entire sequence \cite{gu2023mamba}. Unlike previous state-space models like S4, Mamba performs at a similar level to transformers. Vision Mamba specifically integrates Mamba with visual data processing \cite{zhu2024vision}.

Alongside attacking clean models, we chose the Diffusion Model Adversarial Training Defense on WRN-28-10\cite{wang2023better} - a SOTA defense that integrates adversarial training with a diffusion process to iteratively refine noisy samples. We provide a ResNet-164 trained on CIFAR-10 to serve as a baseline to compare with our defense.


\textbf{Dataset} The original TA paper runs attacks on 200 randomly sampled correctly classified images from ImageNet \cite{wang2022triangle}. We run both TA and TARL on the same 200 images. Furthermore, we test TA and TARL on 1000 classwise-balanced correctly-classified examples for each model in their respective datasets.

\textbf{Evaluation metrics} We count the number of ``successful" adversarial examples as post-attack examples that meet the following criterion:

\begin{enumerate}
    \vspace{-0.15 cm}
    
    \item They are misclassified by the target model. 

    \vspace{-0.25 cm}

    \item The $\mathcal{L}_2$ distance between the adversarial example $x_{adv}$ and the clean example $x$ is no more than the root mean square error $RMSE$ \cite{li2020qeba}: 

    \vspace{-0.7 cm}

\end{enumerate}

\begin{equation}
        C \cdot \sqrt{\frac{1}{w \cdot h \cdot c} \sum_{i=1}^w \sum_{j=1}^h \sum_{k=1}^c (x[i, j, k] - x_{adv}[i, j, k])^2}
\end{equation}

\vspace{-0.10 cm}

where $w$, $h$, and $c$ are the width, height and number of channels of the input image respectively. We vary $C$ as a constant from $0.01$, $0.05$ to $0.1$. 

We refer to the ``attack success rate", abbreviated ASR, as the percentage of successful adversarial examples an attack generates (out of the total number of examples being attacked) under a specific perturbation budget.  


\subsection{Evaluation}

\begin{table}[ht]
\centering
\resizebox{\columnwidth}{!}{%
\begin{tabular}{lcccccccc}
\toprule
& OPT   & SignOPT & HSJA  & QEBA  & GeoDA & Surfree & \begin{tabular}[c]{@{}c@{}}TA \\ (1k query)\end{tabular} & \begin{tabular}[c]{@{}c@{}}TARL \\ (500 query)\end{tabular} \\
\midrule
VGG-16         & 76.0  & 94.0    & 92.5  & 98.5  & 99.0  & 99.5    & 100.0 & \textbf{100.0}    \\
Inception-v3   & 34.0   & 50.5    & 32.5  & 78.5  & 75.5  & 87.5    & 96.5  & \textbf{96.5}   \\
ResNet-18      & 67.0   & 84.5     & 83.0  & 98.0  & 94.5   & 98.5    & 100.0  & \textbf{100.0}   \\
ResNet-101     & 51.5   & 69.0     & 71.5  & 94.0  & 89.5  & 95.5    & \textbf{99.0}  & 97.0   \\
DenseNet-121   & 51.5  & 69.5    & 70.5  & 91.0  & \textbf{100.0} & 97.0    & 99.5    & 99.0 \\
\bottomrule
\end{tabular}%
}
\caption{
\label{table:twohundred}
Results for RMSE = 0.1 across different models on 200 selected images
}
\end{table}


Keeping all hyperparameters consistent except for query number, TARL takes on 500 queries in comparison to the 1,000 queries used for the original TA method, cutting the query efficiency by half. Note that when given a 1,000-query budget, TA's attack success rate outperformed all the previously mentioned popular decision-based attacks \cite{wang2022triangle}.

In our experiments, we first rigorously evaluate the performance of TARL against a range of popular decision-based attacks, using the same experimental setup as TA for consistency, as shown in Table \ref{table:twohundred}. We added an extra column to Table 1 in \cite{wang2022triangle} to effectively compare TARL to other effective decision based boundary attacks including OPT \cite{cheng2018queryefficient}, SignOPT \cite{cheng2019sign}, HopSkipJumpAttack (HSJA) \cite{chen2020hopskipjumpattack}, QEBA \cite{li2020qeba},  GeoDA \cite{rahmati2020geoda}, and Surfree \cite{maho2021surfree}.

In Table \ref{table:twohundred}, we use the same 200 images selected in TA and evaluate across five widely adopted clean image classifiers. To highlight the query efficiency of TARL, we constrain its query budget to 500, whereas all other attacks retain the standard budget of 1000 queries. Despite this significant reduction in query budget, \textbf{TARL outperforms all other methods on three of the five models and achieves comparable results (within 2\% difference in attack success rate, ASR) on the remaining two.}

It is worth noting that, prior to our work, TA had established itself as the state-of-the-art attack method\cite{wang2022triangle}. This further emphasizes the significance of our results, as TARL not only competes with but surpasses TA in terms of query efficiency, and performance in most cases. Given TA’s previous dominance, in subsequent experiments, we focus solely on comparing TA and TARL.

To extend the scope of our evaluation beyond the initial 200 selected images, we apply both TA and TARL to 1000 correctly-classified classwise balanced images for each ImageNet model. We test the performance of these two methods under varying perturbation budgets and further enhance the experimental setup by introducing five additional clean classifiers and even one defense mechanism. This extended evaluation allows us to comprehensively assess the robustness and query efficiency of TARL in a broader and more challenging context.

Tables \ref{table:smallRMSE}, \ref{table:midRMSE}, \ref{table:largeRMSE} and \ref{table:noRMSE} all share the same format. The first column indicates which dataset TA and TARL attacked. The second column lists the victim models. The third and fourth columns report the attack success rate from TA and TARL on each model. 

The fifth column reports the difference between the attack success rate of TARL from TA across each model. A positive difference refers to the percentage of successful adversarial examples that TARL was able to generate that TA was not out of the dataset. A negative difference indicates the percentage of successful adversarial examples TA was able to generate that TARL did not. 

Table \ref{table:smallRMSE} represents our most limiting perturbation setting as it reports the ASR of TA and TARL on all models across all datasets with a perturbation limit of RMSE $C=0.01$. On the ImageNet models, TA outperforms TARL, achieving an average difference of -4.38\%. In perspective, this means TA on-average generates 44 more adversarial examples with an $l_2$ distance no greater than RMSE $C=0.01$ than TARL. 

\begin{table}
\centering
\caption{TA and TARL attack success rate when $\mathcal{L}_2$ distance is no greater than RMSE $C = 0.01$}
\label{table:smallRMSE}
\footnotesize 
\begin{tabular}{@{}lcccc@{}}
\toprule
Dataset & Model & \begin{tabular}[c]{@{}c@{}}TA \\ (1k query)\end{tabular} & \begin{tabular}[c]{@{}c@{}}TARL \\ (500 query)\end{tabular} & Diff. \\
\midrule
ImageNet & VGG-16 & 37.6\% & 33.1\% & -4.5\% \\
ImageNet & Inception-v3 & 46.6\% & 41.0\% & -5.6\% \\
ImageNet & ResNet-18 & 45.4\% & 40.9\% & -4.5\% \\
ImageNet & ResNet-101 & 37.6\% & 32.1\% & -5.5\%\\
ImageNet & DenseNet-121 & 39.4\% & 35.8\% & -3.6\%\\
ImageNet & ViT-L & 17.1\% & 13.6\% & -3.5\% \\
ImageNet & ViT-B & 39.6\% & 32.3\% & -7.3\% \\
ImageNet & BiT-M & 18.4\% & 16.5\% & -1.9\% \\
ImageNet & Mamba & 31.9\% & 28.9\% & -3.0\% \\ 
\midrule 
CIFAR-10 & ResNet-164 & 5.3\% & 5.1\% & -0.2\%\\
CIFAR-10 & Diffusion & 1.7\% & 1.8\% & +0.1\% \\ 
\bottomrule
\end{tabular}
\end{table}

\begin{table}
\centering
\caption{TA and TARL attack success rate when $\mathcal{L}_2$ distance is no greater than RMSE $C = 0.05$}
\label{table:midRMSE}
\footnotesize 
\begin{tabular}{@{}lcccc@{}}
\toprule
Dataset & Model & \begin{tabular}[c]{@{}c@{}}TA \\ (1k query)\end{tabular} & \begin{tabular}[c]{@{}c@{}}TARL \\ (500 query)\end{tabular} & Diff. \\
\midrule
ImageNet & VGG-16 & 92.3\% & 91.3\% & -1.0\% \\
ImageNet & Inception-v3 & 90.8\% & 88.2\% & -2.6\% \\
ImageNet & ResNet-18 & 91.8\% & 89.6\% & -2.2\% \\
ImageNet & ResNet-101 & 82.6\% & 81.4\% & -1.2\% \\
ImageNet & DenseNet-121 & 88.3\% & 84.9\% & -3.4\% \\
ImageNet & ViT-L & 51.1\% & 46.9\% & -4.2\% \\
ImageNet & ViT-B & 86.6\% & 82.8\% & -3.8\% \\
ImageNet & BiT-M & 53.6\% & 48.9\% & -4.7\% \\
ImageNet & Mamba & 72.2\% & 67.7\% & -4.5\% \\
\midrule
CIFAR-10 & ResNet-164 & 55.6\% & 54.8\% & -0.8\% \\
CIFAR-10 & Diffusion & 20.3\% & 19.5\% & -0.8\% \\  
\bottomrule
\end{tabular}
\end{table}



\begin{table}
\centering
\caption{TA and TARL attack success rate when $\mathcal{L}_2$ distance is no greater than RMSE $C = 0.1$}
\label{table:largeRMSE}
\footnotesize 
\begin{tabular}{@{}lcccc@{}}
\toprule
Dataset & Model & \begin{tabular}[c]{@{}c@{}}TA \\ (1k query)\end{tabular} & \begin{tabular}[c]{@{}c@{}}TARL \\ (500 query)\end{tabular} & Diff. \\
\midrule
ImageNet & VGG-16 & 98.8\% & 98.9\% & +0.1\% \\
ImageNet & Inception-v3 & 99.3\% & 99.4\% & +0.1\% \\
ImageNet & ResNet-18 & 99.0\% & 99.3\% & +0.3\% \\
ImageNet & ResNet-101 & 97.4\% & 97.5\% & +0.1\% \\
ImageNet & DenseNet-121 & 97.5\% & 98.4\% & +0.9\% \\
ImageNet & ViT-L & 79.1\% & 76.1\% & -3.0\%\\
ImageNet & ViT-B & 98.1\% & 97.7\% & -0.4\% \\
ImageNet & BiT-M & 75.2\% & 70.7\% & -4.5\% \\
ImageNet & Mamba & 90.9\% & 87.1\% & -3.8\%\\ 
\midrule
CIFAR-10 & ResNet-164 & 88.4\% & 88.3\% & -0.1\% \\
CIFAR-10 & Diffusion & 63.1\% & 61.0\% & -2.1\% \\ 
\bottomrule
\end{tabular}
\end{table}


\begin{table}
\centering
\caption{TA and TARL attack success rate with RMSE $< 0.5$}
\label{table:noRMSE}
\footnotesize 
\begin{tabular}{@{}lcccc@{}}
\toprule
Dataset & Model & \begin{tabular}[c]{@{}c@{}}TA \\ (1k query)\end{tabular} & \begin{tabular}[c]{@{}c@{}}TARL \\ (500 query)\end{tabular} & Diff. \\
\midrule
ImageNet & VGG-16 & 99.2\% & 99.8\% & +0.6\% \\
ImageNet & Inception-v3 & 99.3\% & 99.9\% & +0.6\%\\
ImageNet & ResNet-18 & 99.4\% & 99.6\% & +0.2\% \\
ImageNet & ResNet-101 & 98.9\% & 99.7\% & +0.8\% \\
ImageNet & DenseNet-121 & 99.3\% & 99.9\% & +0.6\% \\
ImageNet & ViT-L & 99.0\% & 99.9\% & +0.9\% \\
ImageNet & ViT-B & 99.2 \% & 100\% & +0.8\% \\
ImageNet & BiT-M & 99.4\% & 99.7\% & +0.3\% \\
ImageNet & Mamba & 99.2\% & 100\% & +0.8\% \\ 
\midrule
CIFAR-10 & ResNet-164 & 99.0\% & 99.0\% & $\pm$ 0\% \\
CIFAR-10 & Diffusion& 99.8\% & 99.8\% & $\pm$ 0\% \\ 
\bottomrule
\end{tabular}
\end{table}

Table \ref{table:midRMSE} reports ASR when the perturbation limit is RMSE $C=0.05$. Here, the ASR of both TA and TARL is more than twice as large as the ASR for each entry in Table \ref{table:smallRMSE}. On the ImageNet models, the margin between TA and TARL is slimmer. The average difference between TARL from TA across all ImageNet models is -3.07\%.

\begin{figure}
  \centering
  \includegraphics[scale=0.5, width=6cm]{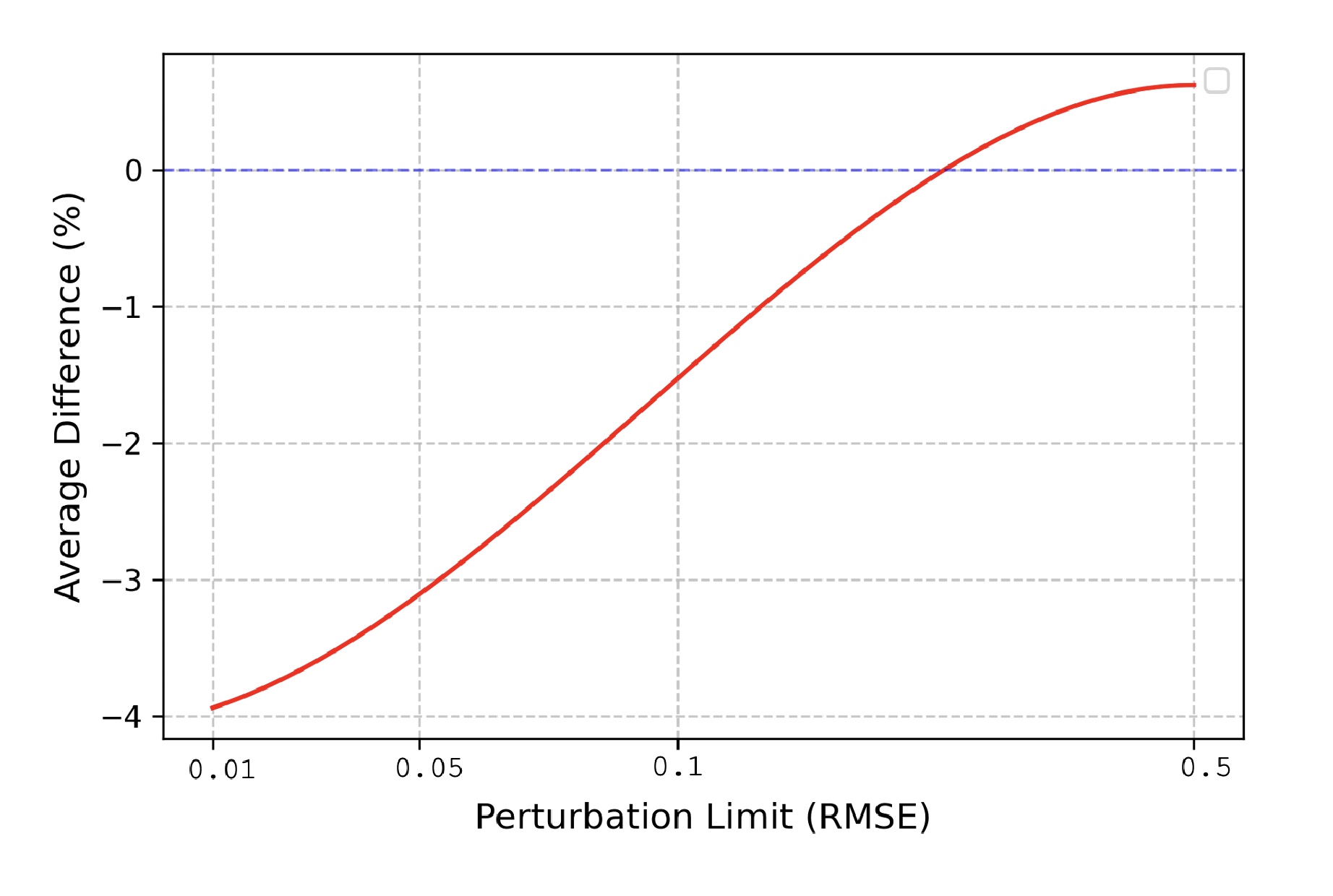}
  \caption{Correlation between RMSE budgets and performance difference between TA and TARL}
  \label{fig:diffAvg}
\end{figure}

Table \ref{table:largeRMSE} reports ASR with RMSE $C=0.1$. Here, most ASR for both TA and TARL are above $80\%$, highlighting the efficiency of the triangle attack methodology. The average difference between TARL from TA on the ImageNet models is -1.13\%. We see a trend: \textbf{when the constant $C$ in RMSE increases, the average difference decreases.} We further demonstrate this trend in figure \ref{fig:diffAvg}.

This highlights a key advantage of TARL: with very slight increases in perturbation budget, TARL maintains a competitive ASR while significantly reducing the number of queries required. This trade-off makes TARL an excellent option in scenarios where time constraints or query budgets are a concern, as it provides a small decrease in ASR for a much larger gain in query efficiency.

Across all tables with a perturbation limit \ref{table:smallRMSE}, \ref{table:midRMSE} and \ref{table:largeRMSE}, on CIFAR-10 both TA and TARL achieve a higher ASR on the base ResNet-164 than the Diffusion model. This shows that defenses are more robust against adversarial examples with a set perturbation limit, as in Table \ref{table:noRMSE}, the Diffusion model performs worse against TA and TARL than the ResNet-164.

These results gives way to an interesting observation: both TA and TARL perform substantially worse on the ViT-L and BiT-M models. Without the ViT-L or BiT-M, the average difference in Table \ref{table:largeRMSE} becomes only -0.3\%. We acknowledge there is future work in improving both TA and TARL to work against attention and transferability-based models.  


Table \ref{table:noRMSE} reports ASR when we run TA and TARL with RMSE $=0.5$, our maximal perturbation limit. Here, \textbf{TARL outperforms TA for every ImageNet model}, producing an average difference of +0.62\% (6 more sucessful adversarial examples). 

In total, TARL performs on average within a 4\% margin of TA, a SOTA black-box attack, with half the query budget. This difference in performance shrinks with a larger perturbation budget as without a perturbation limit, TARL outperforms TA on every model. This highlights the effectiveness of the our revised Triangle Attack method(TARL) across different victim models, making it a promising approach for generating adversarial examples in a query-efficient manner. 

This result underscores the trade-off between perturbation budgets and query efficiency. Even under tight perturbation constraints($C=0.5$), TARL demonstrates its strength by achieving highest attack success rates while significantly reducing query usage. For those seeking a query-efficient attack that remains robust under strict perturbation limits without compromising performance, TARL presents a highly effective solution. Its ability to deliver superior attack success rates with tighter budgets further highlights its practicality in scenarios where both query efficiency and robustness are crucial.

\section{Conclusion}
Our work advances query-efficient decision-based attacks by addressing the theoretical limitations of the state-of-the-art Triangle Attack (TA) and introducing TARL, a more efficient and theoretically sound method. 

Furthermore, TARL reduces the number of queries by half while maintaining nearly identical attack success rates. This achievement is particularly important for real-world applications where query efficiency is crucial, with limited resources or strict time constraints.

Moreover, we expand the scope of previous work by extending the evaluation beyond 200 selected images to the entire ImageNet dataset. We also incorporated additional classifiers and a defense mechanism, providing a comprehensive assessment of TA and TARL under different perturbation budgets. This extended experimental setup highlights TARL’s superior query efficiency and robust performance, establishing it as a new benchmark in decision-based attacks.


{\small
\bibliographystyle{ieee_fullname}
\bibliography{mybibfile}
}

\end{document}